\def\year{2020}\relax
\documentclass[letterpaper]{article} % DO NOT CHANGE THIS
\usepackage{aaai20}  % DO NOT CHANGE THIS
\usepackage{times}  % DO NOT CHANGE THIS
\usepackage{helvet} % DO NOT CHANGE THIS
\usepackage{courier}  % DO NOT CHANGE THIS
\usepackage[hyphens]{url}  % DO NOT CHANGE THIS
\usepackage{graphicx} % DO NOT CHANGE THIS
\usepackage{graphicx}  % DO NOT CHANGE THIS
\usepackage{times}
\usepackage{amsmath}
\usepackage{amssymb}
\usepackage{scalerel}
\usepackage[dvipsnames]{xcolor}
\usepackage{multirow}
\usepackage{subfigure}
\usepackage{breqn}
\title{Visual Relationship Detection with Low Rank Non-Negative \\ Tensor Decomposition}
\author{Mohammed Haroon Dupty\textsuperscript{}\thanks{corresponding author}, Zhen Zhang, Wee Sun Lee\\
School of Computing, National University of Singapore\\
\{dmharoon, leews\}@comp.nus.edu.sg, zhen@zzhang.org
}
 
\begin{document}

\maketitle

\begin{abstract}
We address the problem of Visual Relationship Detection (VRD) which  aims to describe the relationships between pairs of objects in the form of triplets of (\textit{subject, predicate, object}). We observe that given a pair of bounding box proposals, objects often participate in multiple relations implying the distribution of triplets is multimodal. We leverage the strong correlations within triplets to learn the joint distribution of triplet variables conditioned on the image and the bounding box proposals, doing away with the hitherto used independent distribution of triplets. To make learning the triplet joint distribution feasible, we introduce a novel technique of learning conditional triplet distributions in the form of their normalized low rank non-negative tensor decompositions. Normalized tensor decompositions take form of mixture distributions of discrete variables and thus are able to capture multimodality. This allows us to efficiently learn higher order discrete multimodal distributions and at the same time keep the parameter size manageable.
We further model the probability of selecting an object proposal pair and include a relation triplet prior in our model. We show that each part of the model improves performance and the combination outperforms state-of-the-art score on the Visual Genome (VG) and Visual Relationship Detection (VRD) datasets.

%for annotation to handle missing annotations problem
 
%Each rank-1 component of the mixture can represent a single mode and the mixture can represent modes atmost the number of components, thus capturing the multimodality.

\end{abstract}

\section{Introduction}
Object detection is a central problem in computer vision. Recent deep learning approaches~\cite{ren2015faster,girshick2014rich} have made long strides in the task of object detection. %But
%% Modified by Zhen
However, 
real world images often involve multiple objects that interact with each other.
% Just being able to recognize and localize objects is not sufficient to understand such images. 
Much can be said about the image if we can reason object interactions with each other in addition to detection. Reasoning about the relationships objects participate in provides a powerful method for capturing mid-level information that is useful for computer vision tasks; for example, in image captioning, richer captions can be generated if objects as well as the relationships between objects in the image is provided. The task of Visual Relationship Detection aims to recognize and localize the objects along with predicting the relationship that pairs of objects participate in. 

% \begin{figure}[t]
% \centering
% %\fbox{\rule{0pt}{2in} \rule{0.9\linewidth}{0pt}}
% %\includegraphics[width=0.9\columnwidth]{motivation.pdf}
% \includegraphics[scale=0.6]{motivation-1.pdf}
% \caption{Visual relationships are defined with triplets of (subject - predicate - object). Multiple predicates may exist between a pair of box proposals suggesting the distribution of triplets given box proposals is often multimodal. We represent this conditional triplet distribution as normalized low rank tensor decomposition. This representation takes form of mixture distribution of triplet variables thus capturing the multimodality while being efficiently learnable.
% } 
% \label{fig:motivation}
% \label{fig:onecol}
% \end{figure}
\begin{figure}[t]
\centering
\subfigure{\label{subfig:a}\includegraphics[width=0.4\columnwidth]{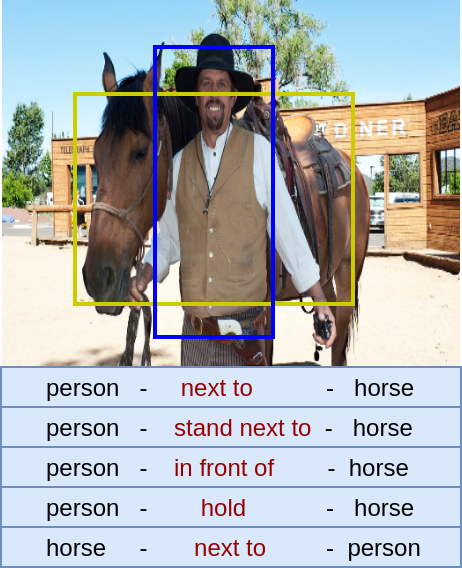}}
\subfigure{\label{subfig:b}\includegraphics[width=0.4\columnwidth]{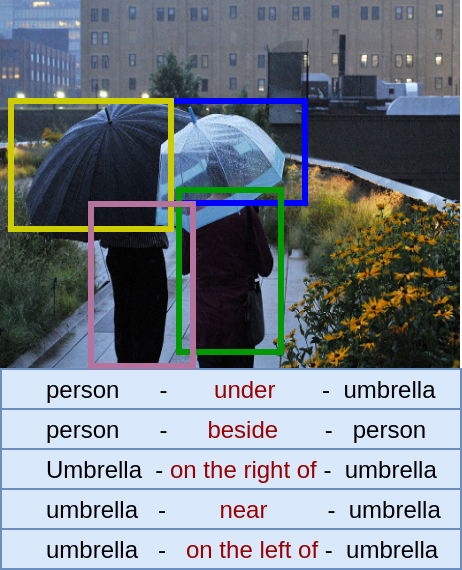}}
\caption{Visual relationships are defined with triplets of (subject - predicate - object). Multiple predicates may exist between a pair of box proposals suggesting the distribution of triplets given box proposals is often multimodal. We represent this conditional triplet distribution as normalized low rank tensor decomposition. This representation takes form of mixture distribution of triplet variables thus capturing the multimodality while being efficiently learnable.
} 
\label{fig:motivation}
\label{fig:onecol}
\end{figure}

State-of-the-art methods for visual relationship detection ~\cite{dai2017detecting,xu2017scene,liang2017deep,zhu2018deep} solve the problem in two stages: an objects proposal stage that uses object detection to propose a set of object bounding box proposals that may participate in relations within the image, and a relation recognition stage that outputs a set of possible relation triplets given the set of bounding box proposals provided by the objects proposal stage. In this paper, we focus on the relation recognition stage. Our main observation is that multiple relations often exist between a pair of box proposals in an image: for example, in Figure~\ref{fig:motivation}, the relations $<$\emph{person - next to - horse}$>$, $<$\emph{person - in front of - horse}$>$ and $<$\emph{person - hold - horse}$>$ are all valid. This suggests that the distribution of triplets, given a pair of objects, is often multimodal with multiple valid relations occurring between a pair of objects. Any learning model with a single separate output for object and predicate classes cannot represent such a multimodal distribution.

\begin{figure}[t]
\begin{center}
   \includegraphics[width=0.5\linewidth]{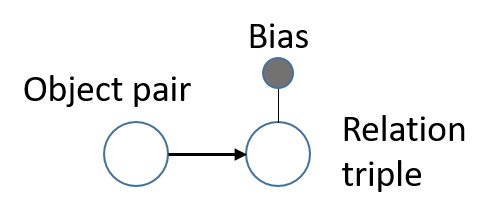}
\end{center}
   \caption{Graphical model representing the relation recognition model. Given an input image and its bounding boxes, a pair of bounding boxes is selected by the annotator and then annotated with a relation triplet. The relation triplet node is augmented by a single node potential representing the bias caused by the prevalence of each triplet in the dataset.}
\label{fig:crf}
\end{figure}

In this work, we use a neural network to model the probability of a triplet $p(t|b,I)$ where $t=(x_s,x_p,x_o)$ consists of the subject, predicate and object labels, $b=(b_s,b_o)$ is the pair of box proposals and $I$ is the image. However, providing a separate output for each triplet combination requires a large number of outputs and a correspondingly large number of parameters in the network; learning such a network would require a large training set. Instead of providing a separate output for each triplet, our neural network outputs a low-rank non-negative tensor decomposition, given each image and pair proposal as input. This limits the number of outputs to be proportional to the sum (instead of product) of the number of values that each variable in the triplet can take. Further, tensor decomposition structure enables computing gradient of loss without construction of full tensor which makes training efficient.%with minimal increase in training time.

The frequency of appearance of the triplets, independent of the proposed bounding box pair, provide a useful bias for improving performance. To incorporate information about the unconditional frequency of appearance of each triplet, we multiply each conditional output with a bias where the bias is represented using a three dimensional non-negative tensor of triplet frequencies. This gives a simple conditional random field representation for the conditional distribution of triplets $ p(t|b) = \psi_c(t|b)\psi_u(t)$ where $\psi_c$ is the conditional potential function represented using a neural network that outputs low-rank non-negative decomposition, and $\psi_u$ is the unconditional potential function to capture the frequency bias, represented using a three dimensional tensor. 

In addition to multi-modality, previous studies \cite{lu2016visual,krishna2017visual} have shown that there exists missing annotation problem in all visual relationship datasets, i.e. a relationship is annotated in certain examples and the same relationship is not annotated in other examples though it exists; for example, it is more likely that pairs of boxes close to each other catch the attention of the annotator than pairs which are farther apart. We model that with probability of a pair of boxes being selected by the annotator for annotation. 

The complete process of generating the relation triplets given an image and a set of detected boxes can be represented with a probabilistic model 
%% Modified by Zhen
shown in Figure~\ref{fig:crf}. In summary, we make following contributions:
%We summarize our main contributions:
%Our main contributions can be summarized as:
\begin{itemize}
    \item We propose a novel way of learning higher order triplet  distributions in the form of their low rank tensor decompositions. This representation takes form of the mixture of rank-1 tensors and thereby is able to represent multimodal distributions.%the representation power required to capture multimodal distributions,
    \item Our formulation enables efficient computation of gradient of the log likelihood and errors can be backpropagated without forming the higher order tensor.
    \item We further augment the conditional triplet distribution with relation frequency prior and probability of annotating a object proposal pair which together outperform the state-of-the-art score on visual relationship detection.
\end{itemize}

\section{Related work}
We need compact form representation of joint distribution of triplets and tensor decompositions are a natural alternative to represent such higher-order functions. There has been a recent surge in using tensor decompositions in various machine learning problems~\cite{anandkumar2014tensor}, ~\cite{janzamin2015beating}, ~\cite{wrigley2017tensor}. 
%Tensor decompositions offer compact representation while satisfing uniqueness properties i.e. for CP-decomposition, the tensor decomposition is unique for a given tensor. 
Low rank representation has been shown to perform reasonably well particularly when the size of the exact model is large. 
%Tensor decompositions have been found particularly useful in the field of graphical models. Recently, tensor decomposition was used for learning latent variable models by representing the their lower order moments as tensor decompositions and using the method of moments to infer the latent variables ~\cite{anandkumar2014tensor}. 
%The closest problem to relationship detection is scene graph generation in which there exists only one predicate between pairs of objects where tensor decompositions have already found there way. 
For the task of visual relationship detection in ~\cite{jae2018tensorize}, the empirical distribution of visual relationships in the dataset was approximated with low rank Tucker decomposition and used as a prior for regularization during learning. 
In contrast to this work, we use tensor decomposition to represent the conditional probability distribution of the relation triplets. 
%Closest to our work is ~\cite{wrigley2017tensor}, who represented the clique potentials of the factor graph as a mixture of there low rank tensor decompositions. With this representation, they were able to run inference algorithm even for large tree-width graphs.

Over the last few years, a number of different approaches have been proposed for the task of recognizing relationships from the image. Most of these methods can be divided into three broad categories. One line of work uses structured prediction techniques by message passing among the three triplet variables~\cite{dai2017detecting,xu2017scene,liang2017deep,zhu2018deep}. Structured prediction techniques are mainly useful as the predicate distribution conditioned on the object labels is highly predictive as was shown in~\cite{dai2017detecting}. They take into account within triplet dependencies by message passing among object and predicate labels.

Another line of work introduces extra information either in the form of word vector embeddings of the object labels or use knowledge from a large corpus~\cite{lu2016visual,yu2017visual,zhang2017visual,zhang2019large}. Learning from a large external knowledge has been shown to be an effective strategy to tackle the missing data problem~\cite{yu2017visual}.  Other prevalent methods ~\cite{lu2016visual,zhang2019large,zhang2017visual,yu2017visual} have used a combination of visual and linguistic features to detect relationships between objects and showed the utility of word vector embeddings. With much work done, it is now well established that the use of word vectors is substantially helpful in recognizing relationships ~\cite{lu2016visual,yu2017visual,zhang2017visual,zhang2019large}.

A third line of work uses rank-based loss functions to encourage similar relations to be close to each other in the learnt feature space~\cite{liang2018visual,zhang2019large}. Most recently ~\cite{zhang2019large} used triplet loss to match the visual and semantic features in a projected shared space for better discriminative power.  

In this work we look only from the visual perspective without any sort of external information. Instead, we try to capture the multimodal properties of the triplet distribution and to model the generative annotation process. 
 
\section{Preliminaries: Tensor decomposition}
Tensors are generalizations of matrices to higher dimensions and hence a tensor can be called a multidimensional array. A order-$d$ tensor $T$ is an element in $\mathcal{R}^{N_1\times N_2\cdots\times N_d}$ with $N_k$ possible values in $k^{th}$ dimension. 
Analogous to SVD in matrices, tensors can be represented in succinct form with tensor decompositions. In CANDECOMP / PARAFAC (CP) decomposition, any tensor $T$ can be represented as a linear combination of outer products of vectors as
\begin{equation}
T = \sum_{r=1}^R w_r \phi_{r,1} \otimes \phi_{r,2} \otimes \cdots \otimes \phi_{r,d}
\label{eq:tensor}
\end{equation}
where $\otimes$ is the outer product operator, each $\phi_{r,k}$ is a vector in $R^{N_k}$ for $k \in \{1,2,\dots,d\}$ and the term $\phi_{r,1} \otimes \phi_{r,2} \otimes \cdots \otimes \phi_{r,d}$ is a rank-1 tensor. $w_r$ is a scalar co-efficient which can be absorbed in one of the vectors $\phi_{r,k}$. Tensor value at index ($i_1,i_2\dots,i_d$) is given by $\sum_{r=1}^R w_r \phi_{r,1}^{i_1} \phi_{r,2}^{i_2} \cdots \phi_{r,d}^{i_d}$. The smallest $R$ for which an exact $R$-term decomposition exists is the rank of tensor $T$ and the decomposition (\ref{eq:tensor}) is its $R$-rank approximation of the tensor. With this compact representation a tensor $T$ with ${N_1\times N_2\cdots\times N_d}$ entries can be represented with $R$ vectors for each variable in $T$ i.e. with $R(N_1+N_2+\dots+N_d)$ entries. More information about tensor decompositions can be found in ~\cite{kolda2009tensor,rabanser2017introduction}. 

With low rank assumption, we can represent any probability distribution of multiple variables in tensor form of (\ref{eq:tensor}), provided we constrain the values of $T$ to be non-negative and normalize it such that sum of all entries of $T$ is $1$. We call such a normalized non-negative tensor a \textbf{\emph{probability tensor}}. Note that the form of the \textbf{\emph{probability tensor}} in (\ref{eq:tensor}) is more like mixture distribution of discrete variables and thus is suitable for modeling multimodal triplet distribution.

\section{Formulation}
%Given an image, the task of Visual Relationship Detection (VRD) involves extracting a set of visual relationships in the form of (subject - predicate - object) triplets and localizing the objects as bounding boxes in the image. 
Like other recent works~\cite{dai2017detecting,liang2017deep,zhu2018deep}, we treat task of visual relationship detection with a two stage pipeline where the boxes are given by a separately trained object detector, Faster-RCNN~\cite{ren2015faster} and a classifier predicts the relationships between each pair of the boxes including a null relation for box pair that do not participate in a relation.% In this work, we use Faster-RCNN~\cite{ren2015faster} for object detection and focus on relationship recognition by asking given a set of detected box proposals, how best to recognize relationships between pairs of box proposals. 

Formally given an image $I$ with $N$ objects represented by rectangular bounding box proposals, we have $N(N-1)$ relations between each pair of the objects. For each pair of subject and object box proposal $(B_s=b_s, B_o=b_o)$, subject label $X_s=x_s$, predicate label $X_p=x_p$ and object label $X_o=x_o$ form a triplet instance.  

\begin{figure*}
\centering
\includegraphics[width=0.85\textwidth]{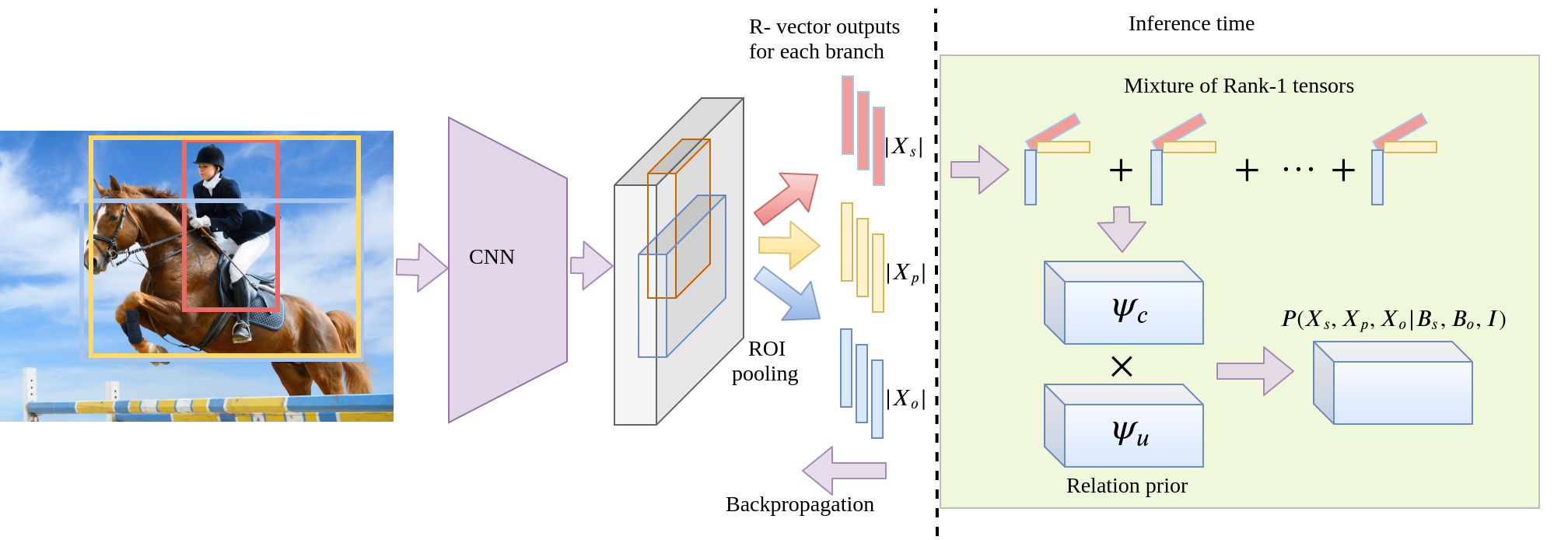}
\caption{An image is input to a neural network (VGG16) to produce an intermediate feature map. For each pair of boxes from the detector, the corresponding features are ROI-pooled and fed through three separate branches of fully connected layers each for subject, predicate and object. Each branch outputs a set of $R$ vectors which together form the mixture of independent triplet distributions($\psi_c$) capturing the multimodal distribution. $\psi_c$ is multiplied with unconditional relation prior $\psi_u$ constructed from the training set to give $P(X_s,X_p,X_o|B_s,B_o,I)$. During training, construction of $\psi_c$ is not required and errors can be backpropagated from the set of $R$ vectors.}
\label{fig:main_model}
\end{figure*}

\subsection{Conditional triplet distribution} 
We use two sources of information for modeling the conditional triplet distribution $P(X_s, X_p, X_o| B_s, B_o, I)$. The first source is information available in the image and bounding box pair. The second is the prior distribution of triplets, not conditioned on the image. We represent the conditional triplet distribution as a product of the two potential functions  $P(X_s, X_p, X_o| B_s, B_o, I)=\psi_c(X_s, X_p, X_o| B_s, B_o, I) \cdot \psi_u(X_s, X_p, X_o) $ to give a simple conditional random field.

$\psi_u(X_s, X_p, X_o)$ is constructed from the training set. It is a order-3 tensor representing the number of times each triplet occurs in the dataset. It serves as a frequency bias towards most frequently occurring relations. $\psi_u(X_s, X_p, X_o) $ $\in \mathbb{R}^{|X_o| \times |X_p| \times |X_o|}$ (for $|X_o|$ object and $|X_p|$ predicate classes). The value at $\psi_u(i, j, k)$ is the number of times triplet $(x_s^i,x_p^j,x_o^k)$ occurs in the training set where $x_s^i$ is the $i^{th}$ subject class, $x_p^j$ is the $j^{th}$ predicate class and $x_o^k$ is the $k^{th}$ object class. It turns out that, only $\sim 1\%$ of the tensor constructed from the training set has non-zero entries. Multiplication with $\psi_c$ will cause all unseen relation triplets to vanish. To address this issue, we smoothen $\psi_u$ by adding 1 to all entries. We then normalize $\psi_u$ by dividing with its sum to get prior probability of occurrence of all relation triplets.

$\psi_c(X_s, X_p, X_o| B_s, B_o, I) \in \mathbb{R}^{|X_o|\times |X_p| \times |X_o|}$ is also an order-3 tensor i.e. given an image and a pair of boxes, $\psi_c$ gives probability of triplet labels. We assume that it is well approximated with a low rank tensor. Our assumption stems from the observation that out of all possible relationships,  only certain specific relationships tend to occur frequently given the underlying objects. The sparsity of prior $\psi_u$ further strengthens our low rank assumption of the tensor. As the rank of any tensor cannot exceed the number of non-zero entries in the tensor, assuming that the tensor is low rank appears to be reasonable.  Consequently, we use a mixture of independent rank-1 tensors to represent $\psi_c(X_s, X_p, X_o|B_s, B_o, I)$, allowing us to effectively capture richer multimodal triplet distributions with a reasonably compact model. We represent the  order-$3$ tensor $\psi_c(X_s, X_p, X_o| B_s, B_o, I)$, in CP-decomposition form as
\begin{dmath}
	\psi_c(X_s, X_p, X_o| B_s, B_o, I) =  \sum_{r=1}^{R} w_r  \phi_{rs}(X_s|B_s, B_o, I)\otimes \phi_{rp}(X_p|B_s, B_o, I)\otimes\phi_{ro}(X_o|B_s, B_o, I)).
	\label{tensor_pot}
\end{dmath}

For notational convenience, we omit the conditioning on the image and bounding boxes in $\phi_{ra}$ from here onwards, where $\phi_{ra}(X_a)$ is the $r^{th}$ vector of the variable $X_a$ for each $a\in \{s,p,o\}$. We parameterize $\psi_c(X_s, X_p, X_o| B_s, B_o, I)$ with a deep neural network and learn it from the data. Given an image and a pair of bounding boxes, the neural network outputs a set of $R$ vectors $s_{a,r}$ for $a\in \{s,p,o\}$. Since potential functions are required to be non-negative, we represent $\phi_{ra}(i) = e^{s^i_{a,r}}$ so that the tensor decomposition is non-negative. Then, we normalize the output tensor by dividing with the sum of all tensor entries to make it a \textbf{\emph{probability tensor}}.

Without loss of generality, we assume the weights $w_r$ in eqn (\ref{tensor_pot}) can be absorbed into the vectors $\phi_{ra}$ and hence there no need for their separate representation. Our representation reduces the number of outputs required to represent the function from $|X_s|\times|X_p|\times|X_o|$ to $R(|X_s|+|X_p|+|X_o|)$. A smaller sized output corresponds to a smaller number of model parameters, making it possible to learn the model from less data. Finally, prior $\psi_u(X_s, X_p, X_o)$ is multiplied elementwise with $\psi_c(X_s, X_p, X_o| B_s, B_o, I)$ to get the conditional triplet distribution $P(X_s, X_p, X_o| B_s, B_o, I) =\psi_c(X_s, X_p, X_o| B_s, B_o, I)\psi_u(X_s, X_p, X_o)$. Note that due to low rank assumption of the $\psi_c$, the network may output spurious triplets which may not be seen in the training set. Probablities of such spurious triplets are pushed down by multiplication with the prior. 
%Normalization allows more efficient piecewise learning of the conditional random field \cite{lin2016efficient}. Marginalization, which is one of the key operation in piecewise learning, can be performed efficiently when we have CP-representation. 

%Out of $|X_s|\times|X_p|\times|X_o|$ possible relations, only an extremely small fraction of around $1\%$ actually appears in the dataset. This observation has been exploited in multiple works ~\cite{yu2017visual,jae2018tensorize,baier2017improving}. In ~\cite{jae2018tensorize}, a rank-R approximation is found and used a regularizer, encouraging the model to mimic the low rank structure of the prior distribution.

\subsection{Training Loss}
We learn the triplet distribution $P(X_s,X_p,X_o|B_s,B_o,I)$ $=$ $\psi_c(X_s,X_p,X_o|B_s,B_o,I)$ $\psi_u(X_s, X_p,X_o)$ using a deep neural network. As both $\psi_c$ and $\psi_u$ are normalized to sum to 1, they both can be trained separately by treating $\psi_c(X_s,X_p,X_o|B_s,B_o,I)$ as a conditional distribution and $\psi_u(X_s, X_p, X_o)$ as a prior distribution. This type of learning is often called piecewise training of the conditional random field \cite{lin2016efficient}. The prior distribution is learned simply by computing the frequencies of the triplets in the training set.

For conditional distribution $\psi_c(X_s, X_p, X_o| B_s, B_o, I)$, we use a neural network that outputs a set of vectors that correspond to the tensor decomposition, given an image and pair of bounding boxes. The network outputs a set of $R$ vectors each for subject, predicate and object for a total of $3R$ vectors. Each of these  $R$ vectors are indexed by $i,j,k$ for subject, predicate and object categories respectively.  Let $s_{a,r}$ be the final layer output of the network for each $a\in \{s,p,o\}$. We exponentiate $s_{a,r}$ and then normalize to make it a \textbf{\emph{probability tensor $y$}}. $(i,j,k)^{th}$ entry of  $y$ can be computed as
\begin{dmath}
    y^{i,j,k} = \frac{\sum_{r=1}^{R} e^{s_{s,r}^i} e^{s_{p,r}^j}  e^{s_{o,r}^k}}{\sum_{l,m,n} \sum_{r=1}^{R} e^{s_{s,r}^i} e^{s_{p,r}^j}  e^{s_{o,r}^k}}.
\end{dmath}
For training such a network, we use cross-entropy loss as it is a classification problem. But unlike the usual case, the loss is computed between tensors instead of vectors. If $t$ representss the target one-hot tensor (tensor that is zero everywhere, except at a single tuple index which has value 1, representing the indicator function of the tuple) and $y$ output \emph{probability tensor}, the cross-entropy loss function is
\begin{dmath}
    L = -\sum_{i,j,k} t^{i,j,k}\log(y^{i,j,k}).
    \label{eq:crs_entropy}
\end{dmath}
A simple brute force method to compute eqn (\ref{eq:crs_entropy}) is to fully construct tensor $y$ from the network output and then compute loss. But this would significantly slow down training. Instead we compute the derivative of loss with respect to final layer output $s_a$ directly without constructing probability tensor $y$. %Our derivation is possible given the outer summation over all variables of the tensor can be pushed inside and evaluated only on single variables. We now describe how to compute derivative of loss without constructing full tensor $y$.

Consider the derivative of the loss w.r.t $s_{s,r'}^i$ where subscript $s$ indicates subject variable $X_s$, $r'$ is one of the $R$ vectors and $i$ is the $i^{th}$ index out of $|X_s|$ indices. For an observed tuple $(i',j',k')$, the derivative of the Loss $L$ with respect to $s_{s,r'}^i$ is given by:
\[ 
\frac{\partial L}{\partial s_{s,r'}^i} =
    \begin{cases}
     \frac{e^{s_{s,r'}^i}\sum_{m'}e^{s_{p,r'}^{m'}}  \sum_{n'}e^{s_{o,r'}^{n'}}}{Z} & i\neq i' \\
     \\
     \frac{e^{s_{s,r'}^i}\sum_{m'}e^{s_{p,r'}^{m'}}  \sum_{n'}e^{s_{o,r'}^{n'}}}{Z} \\ - 
     \frac{e^{s_{s,r'}^i + s_{p,r'}^{j'} + s_{o,r'}^{k'}}}{\sum_{r=1}^{R} e^{s_{s,r}^i + s_{p,r}^{j'} + s_{o,r}^{k'}}}
     & i=i'
    \end{cases}
\]
%where $s_{s,r'}^i$ is the $i^{th}$ network output corresponding to variable $X_s$ and rank $r'$ and 
where 
\begin{dmath}
    Z = \sum_{l,m,n} \sum_{r=1}^{R} e^{s_{s,r}^l + s_{p,r}^m + s_{o,r}^n}
\end{dmath}
is the partition function. 

Naive computation of $Z$ by summing over all entries of tensor $y$ may significantly slow down training. Instead we can compute the partition function efficiently in time linear with class size of each variable $X$, by simply pushing the outer sum $\sum_{l,m,n}$ inside and only evaluate it over the corresponding univariate vectors i.e.
\begin{dmath}
Z = \sum_{r=1}^{R} \sum_l e^{s_{s,r}^l} \sum_m  e^{s_{p,r}^m} \sum_n e^{s_{o,r}^n}. 
\end{dmath}
The derivatives with respect to the predicate and object variable outputs $e^{s_{p,r'}}$ and $e^{s_{o,r'}}$ are computed in a similar way and backpropagated to optimize the network. (derivation shown in supplementary)

\subsection{Modeling missing annotations}
The problem of missing annotations in relationship detection datasets is well known ~\cite{lu2016visual,krishna2017visual}. A relationship is annotated in certain examples and the same relationship may not be annotated in other examples though it exists. We have $N(N-1)$ pairs for $N$ object boxes and only few of them are annotated with relations, rest of the pairs are considered \emph{null}. Some of these \emph{null} pairs may have valid relations but are not annotated. This confuses the model during training as examples with similar features are considered valid as well as \emph{null}. We handle this problem by training $P(X_s, X_p, X_o| B_s, B_o, I)$ with only annotated positive relations. We train a separate binary variable with equal number valid and \emph{null} examples to give $P(X_{sel}|B_s, B_o, I)$, probability of annotation of box pair. At test time, $P(X_{sel}| B_s, B_o, I))$ is multiplied with $P(X_s, X_p, X_o|B_s, B_o, I))$ before final ranking. With this technique, triplet network produces reliable values for valid relations and the unreliable values produced for box-pairs with \emph{null} relation are brought down by multiplication with $P(X_{sel}|B_s, B_o, I)$.

%\textbf{Overall pipeline}
The final scoring function $f_{spo}$ for each of the tuple $(X_s, X_p, X_o, B_s, B_o)$ is given by the full posterior of the relation triplet and pairs of bounding boxes for a given image. 
{\small
\begin{dmath}
    P(B_s,B_o|I) P(X_s, X_p, X_o| B_s, B_o, I) P(X_{sel}|B_s, B_o,I)
\end{dmath}
}
where $P(B_s,B_o|I)$ is from detector output.
\section{Network Architecture}
Figure (\ref{fig:main_model}) shows the workflow of our model. We use VGG16~\cite{simonyan2014very} as backbone of our network initialized with pretrained weights on Visual Genome for detection with freezed conv1\_1-conv5\_3 layers and train it with the ground truth gold proposal boxes. We feed the network with the image to get a global feature map of the image from which subject, predicate and object features are ROI-Align pooled w.r.t their corresponding box regions. The predicate feature is pooled from the union-box of the subject and object boxes. After ROI-pooling, visual features are fed through three separate branches each for subject, object and predicate. 

Parallelly, we include 2-channel spatial binary mask feature of size $2\times64\times64$ as in~\cite{dai2017detecting}. Each channel is a matrix with 1 in bounding box region of object (scaled to size $64\times64$) and 0 everywhere. This feature is passed through 2 convolution layers and then a fully connected layer to get a 512-dim \emph{spatial feature} which is concatenated with ROI-pooled predicate feature. Each of the subject, predicate and object branches has two fully connected layers of size 4096 with final layer output size of $R\times |X_a|$ for $a\in\{s,p,o\}$. %The predicate branch has three fully connected layers. After two FC layers, we concatenate feature vector of predicate with the corresponding features from the subject and object branches along with the $512$-dim mask feature vector. As a result, the predicate branch has one more FC layer than the other two branches. This fused feature vector is fed into the final layer of the predicate branch to output $R\times |X_p|$ sized vector. Note that we do not backpropagate from this concatenation to the object branches as as we found that it brings down the object recognition performance. Identification of the predicate is likely strongly dependent on the object categories, but not vice-versa. 

This ROI-pooled predicate feature concatenated with \emph{spatial feature} also serves as the input for binary classifier for $X_{sel}$ with a single hidden layer of 4096. We first train the triplet network with only annotated positive relationships. We then freeze weights of triplet network and train $X_{sel}$ with equal number of positive and negative examples. 
%This training procedure effectively saves the triplet model from seeing similar examples with both positive and negative labels. For all those negative instance box pairs which are not seen during training, the model produces garbage values and are brought down by the $X_{sel}$ variable for negative classes. 
\begin{table*}[ht]
\centering
\scriptsize
\begin{tabular}{c|cccc|cccc|cccc}
		\hline
		%\multirow{2}{*}{Method} &
		&
		\multicolumn{2}{c}{Relationship } & \multicolumn{2}{c}{Phrase} &
		\multicolumn{4}{c}{Relationship detection} & 
		\multicolumn{4}{c}{Phrase detection} \\
		
		& \multicolumn{4}{c}{mult preds (free k)} & \multicolumn{2}{c}{k=1} &\multicolumn{2}{c}{k=10} & %\multicolumn{2}{c}{(k=70)} &
		\multicolumn{2}{c}{k=1} &\multicolumn{2}{c}{k=10}  \\%\multicolumn{2}{c|}{(k=70)} \\
		
		Recall at & 50 & 100 & 50 & 100 & 50 & 100 & 50 & 100 & 50 & 100 & 50 & 100  \\
		\cline{1-13}
		\textbf{w/ proposals from \cite{lu2016visual}} &&&&&&&&&&&&\\
		Language Cues~\cite{plummer2017phrase} & 16.89 & 20.70 & 15.08 & 18.37 & - & - & 16.89 & 20.70 & - & - & 15.08 & 18.37\\
		VRD~\cite{lu2016visual} & 17.43 & 22.03 & 20.42 & 25.52 & 13.80 & 14.70 & 17.43 & 22.03 & 16.17 & 17.03 & 20.42 & 25.52 \\ 
		
		LargeVRU~\cite{zhang2019large} & 19.18  & 22.64 & 21.69 & 25.92 & 16.08 & 17.07 & 19.18 & 22.64 & 18.32 & 19.78 & 21.69 & 25.92\\ 

		Ours  & \bfseries 24.08 & \bfseries 28.29 & \bfseries	29.17 & \bfseries 34.33 & \bfseries 17.67 & \bfseries 18.64 & \bfseries 24.08 & \bfseries 28.29 & \bfseries 20.80 & \bfseries 22.13 & \bfseries 29.17 & \bfseries 34.33  \\
		\hline
		\textbf{w/ better proposals} &&&&&&&&&&&&\\
		L distilation(Yu et al., 2017) & 22.68 & 31.89 & 26.47 & 29.76 & 19.17 & 21.34 & 22.56 & 29.89 & 23.14 & 24.03 & 26.47 & 29.76 \\
		Zoom-Net~\cite{yin2018zoom} & 21.37 & 27.30 & 29.05 & 37.34 & 18.92 & 21.41 & - & - & 24.82 & 28.09 & - & - \\
		CAI + SCA-M~\cite{yin2018zoom} & 22.34 & 28.52 & 29.64 & 38.39 & 19.54 & 22.39 & - & - & 25.21 & 28.89 & - & - \\
		LargeVRU~\cite{zhang2019large} & 26.98  & 32.63 & \bfseries 32.90 & 39.66 & 23.68 & 26.67 & 26.98 & 32.63 & 28.93 & 32.85 & \bfseries 32.90 &  39.66 \\ 
        MF-URLN~\cite{zhan2019exploring} & 23.9 & 26.8 & 31.5 & 36.1 & 23.9 & \bfseries 26.8 & - & - & \bfseries 31.5 & \bfseries 36.1  & - & - \\
		Ours  & \bfseries 27.09 & \bfseries 34.93 &  32.29 & \bfseries 41.28 & \bfseries 24.20 & 25.87 & \bfseries 27.09 & \bfseries 34.93 & 28.53 & 30.92 &  32.29 & \bfseries 41.28 \\
		\hline
\end{tabular}
\caption{Comparison with state of the art methods on VRD dataset.}
\label{tab:tab_vrd_full}
\end{table*}

\begin{table}[ht]
\centering
\scriptsize
\renewrobustcmd{\bfseries}{\fontseries{b}\selectfont}
\renewrobustcmd{\boldmath}{}
\begin{tabular}{cccccc}
		\hline
		\multirow{2}{*}{Method} & \multicolumn{2}{c}{Relation Detection} &
		\multicolumn{2}{c}{Phrase Detection} \\
		& R@50 & R@100 & R@50 & R@100  \\
		\hline
		VTranseE~\cite{zhang2017visual} & 5.5 & 6.0 & 9.5 & 10.5\\
		PPRFCN~\cite{zhang2017ppr} & 6.0 & 6.9 & 10.6 & 11.1 \\
	    DSL~\cite{zhu2018deep} & 6.8 & 8.0 & 13.1 & 15.6 \\ 
		VSA~\cite{han2018visual} & 6.0 & 6.3 & 9.7 & 10.0\\ 
		MF-URLN~\cite{zhan2019exploring}  & 14.4 & 16.5 & 26.6 & 32.1 \\
        \hline
        Ours (k=1)  & \bfseries 16.74 & \bfseries 18.69 & \bfseries 29.32 & \bfseries 33.42\\
		Ours (free k) & \bfseries 18.52 & \bfseries 21.92 & \bfseries 31.58 & \bfseries 38.07 \\
		\hline
\end{tabular}
\caption{Comparison with state-of-the-art on VG200 dataset}
\label{tab:tab_vg200}
\end{table}

\textbf{Implementation details:} We implement our method on pytorch, a mainstream deep learning library. We set the learning rate to 1e-4 and use SGD as optimizer. Due to the summation in the gradient term, there is an exploding gradient problem. To fix this, we clip the gradient based on the total norm of all the learnable weights. The norm value for gradient clipping is set at $20$. We then train with proposals from the detector. We sample atmost 4 proposals for every ground truth box proposal with IOU overlap of atleast 0.5. %saFor the task of relationship/phrase detection, as we are testing on non-gold proposals on VRD dataset, we augment the the training set with noisy proposals generated by perturbing the gold box proposals. 
All the layers before ROI-pooling are initialized by pretrained weights from the detector. During inference, we filter out the overlapping boxes from the detector by enforcing Non-Maximum Suppression (NMS) constraints with NMS threshold set at 0.7\footnote{https://github.com/dmharoon/VRD-Tensor-Decompostion}.

\textbf{Computation time:} 
With our low-rank tensor formulation of $\psi_c$, we are able to train the triplet network with VRD dataset at 6 min/epoch and VG dataset at 90min/epoch for 7 epochs. Backpropagation on a single image takes 0.21sec on average. Inference for a single image takes around 0.57s with gold proposals. Training time per image is substantially low compared to its inference time as we do not construct full 3D tensor $\psi_c$ during training.

\section{Experiments}
We evaluate our method on the Visual Genome and the Visual Relationship Detection datasets. 

\textbf{VG}: The Visual Genome dataset was released by ~\cite{krishna2017visual}. Unfortunately there is not a single version of Visual Genome that is consistently used by all previous works on this task. To show better performance of our model across splits of VG, we use two different versions in our experiments, \textbf{VG200}~\cite{zhang2017visual} and \textbf{VG150}~\cite{xu2017scene} for comparison with other recent works. \textbf{VG200} has $200$ object and $100$ predicate categories. \textbf{VG150} has $150$ object and $50$ predicate categories. We conduct our ablation studies on VG150.\\
\textbf{VRD:} The VRD dataset was released by ~\cite{lu2016visual} with standard train/test split $4000$ and $1000$ images respectively. There are $100$ object and $70$ predicate categories with $6,672$ unique relationships. On average there are $24.25$ relationships per object category. 

\textbf{Evaluation tasks:}
Consistent with prior works, we report results on two tasks, Relationship Detection and Phrase Detection. In both the tasks we are given an input image and required to output top-50/100 relation triplets with the corresponding bounding boxes for each pair. In \textbf{Relationship detection}, a prediction is considered correct if all three triplet labels (s,p,o) are correctly recognized, and the intersection over union (IOU) between the predicted and the ground-truth boxes is at least 0.5. In \textbf{Phrase detection} the prediction is correct if triplet labels match and IOU of the union of the two predicted boxes with the union of ground-truth boxes is at least 0.5. In line with previous works ~\cite{yu2017visual,zhang2019large}, we consider $k$ relationship predictions per object box pair before taking the top-50/100 predictions per image. We report for $k=1,10$ and \emph{free $k$} ($k$ is cross-validated). For the ablation studies, we fix ground truth boxes as object proposals and report recall on the \textbf{Phrase Classification} task, which masks errors from detector. We report our ablation studies on VG150~\cite{xu2017scene} dataset. 

%Since, our main contribution is in developing a method that works well when there are multiple modes, we report scores on multiple predictions as well as single %prediction per pair of boxes.

\subsection{Results}
\definecolor{orange}{HTML}{F8CECC}
\definecolor{blue}{HTML}{DAE8FC}
\definecolor{green}{HTML}{D5E8D4}
\begin{table}[ht]
	\centering
	\scriptsize
	\renewrobustcmd{\bfseries}{\fontseries{b}\selectfont}
	\renewrobustcmd{\boldmath}{}
	\begin{tabular}{cccccc}
		\hline
		\multirow{2}{*}{Method} & \multicolumn{2}{c}{PhrCls (free k)} &
		\multicolumn{2}{c}{PhrCls (k=1)} \\
		& R@50 & R@100 & R@50 & R@100  \\
		\hline
		Freq-Overlap~\cite{zellers2018neural} & 39.0 & 43.4 & 32.3 & 32.9\\
		Message Passing~\cite{xu2017scene} & 43.4 & 47.2 & 34.6 & 35.4 \\
		Motifnet~\cite{zellers2018neural} &  44.5 & 47.7 & 35.8 & 36.5 \\
		RelDN~\cite{zhang2019graphical} & \bfseries 48.9 & 50.8 & \bfseries 36.8 & 36.8 \\
		Ours &  47.54 & \bfseries 54.69 & 35.60 & \bfseries 37.68\\  \hline
	\end{tabular}
	\caption{Comparison with state-of-the-art on VG150 dataset}
	\label{tab:tab_vg150}
\end{table}
Table~\ref{tab:tab_vrd_full} shows results on VRD dataset. The quality of bounding box proposals from the detector have significant effect on relationship recall results. For a fair comparison, we divide the results in two parts based on the detection box proposals used. We compare previous works which use test set detection proposals from ~\cite{lu2016visual} and also report results with improved proposals from the detector, generated in a manner similar to ~\cite{zhang2019large}. 

Our method shows significant improvement over the prior state-of-the art methods with proposals from ~\cite{lu2016visual}. On relationship detection, there is $5\%$ point increase over the previous best results. These results are equally well translated to the task of phrase detection where our model is able to achieve nearly $8\%$ points improvement. As the proposals used are same, we can infer that the improvement is directly from better relationship recognition. With improved proposals, there is further improvement in the score. 

The results on Visual Genome is shown in Tables~\ref{tab:tab_vg200} and~\ref{tab:tab_vg150}. It is not clear value of $k$ used in prior works for VG200 dataset. Hence we report our results for $k=1$ and free-$k$. Our method performed best in both the tasks in both Recall-50/100. With k=1, we outperform the most recent state-of-the-art on VG200 dataset ~\cite{zhan2019exploring}. by a margin of $2.3\%$. and with cross validated $k$, we achieve 4.5\% improvement over state-of-the-art for Relation detection task at Recall-50. Similarly, there is significant improvement in Phrase Classification for VG150 split. It should be noted that in both datasets, Recall-100 has significantly higher score. This indicates that multimodal distribution is perhaps better captured with our model as our model optimizes to push higher scores for multiple valid relations which is reflected in Recall score with $k>1$.
It should be noted that our method performed very well for the main evaluation metric that we are interested in, where multiple predictions from each bounding box pair are allowed, supporting our claim of capturing multimodal distribution of triplets. 

%This result is without including the dataset prior factor during inference. With prior included, we further get $2$ more points improvement. This shows the effectiveness of the prior information in predicting relationships. Though few techniques previously proposed have tried to use this information~\cite{yu2017visual,baier2017improving}, our principled combination of triplet probabilities from likelihood and prior factor achieves best results on all metrics on VRD dataset. 
 
%The reason for such improvement is that VRD dataset contains multiple relations between the same pairs which has been well captured by out mixture model and augmented with the prior factor.
 
\subsection{Ablation Study and Analysis of Results}
The most distinct part of our model is the mixture distribution model. As the representative power of mixture models increases with increasing mixing components, we first evaluate our model with varying Rank of tensor decomposition or number of mixing components of the model. To evaluate the effectiveness of each of these components, we report \textbf{phrase classification} results on VRD and VG150 datasets.  Phrase classification isolates the factor of object localization accuracy by using ground truth boxes, meaning that it focuses more on the relationship recognition ability of a model. The first 5 models are tested without the dataset prior or $X_{sel}$ variable to analyze the relative gains of increasing the number of mixing components. We further evaluate the effect of $X_{sel}$ variable and dataset prior factor with addition to Rank-5 model.
\begin{figure}[t]
\centering %scale=0.4
\includegraphics[width=0.9\columnwidth]{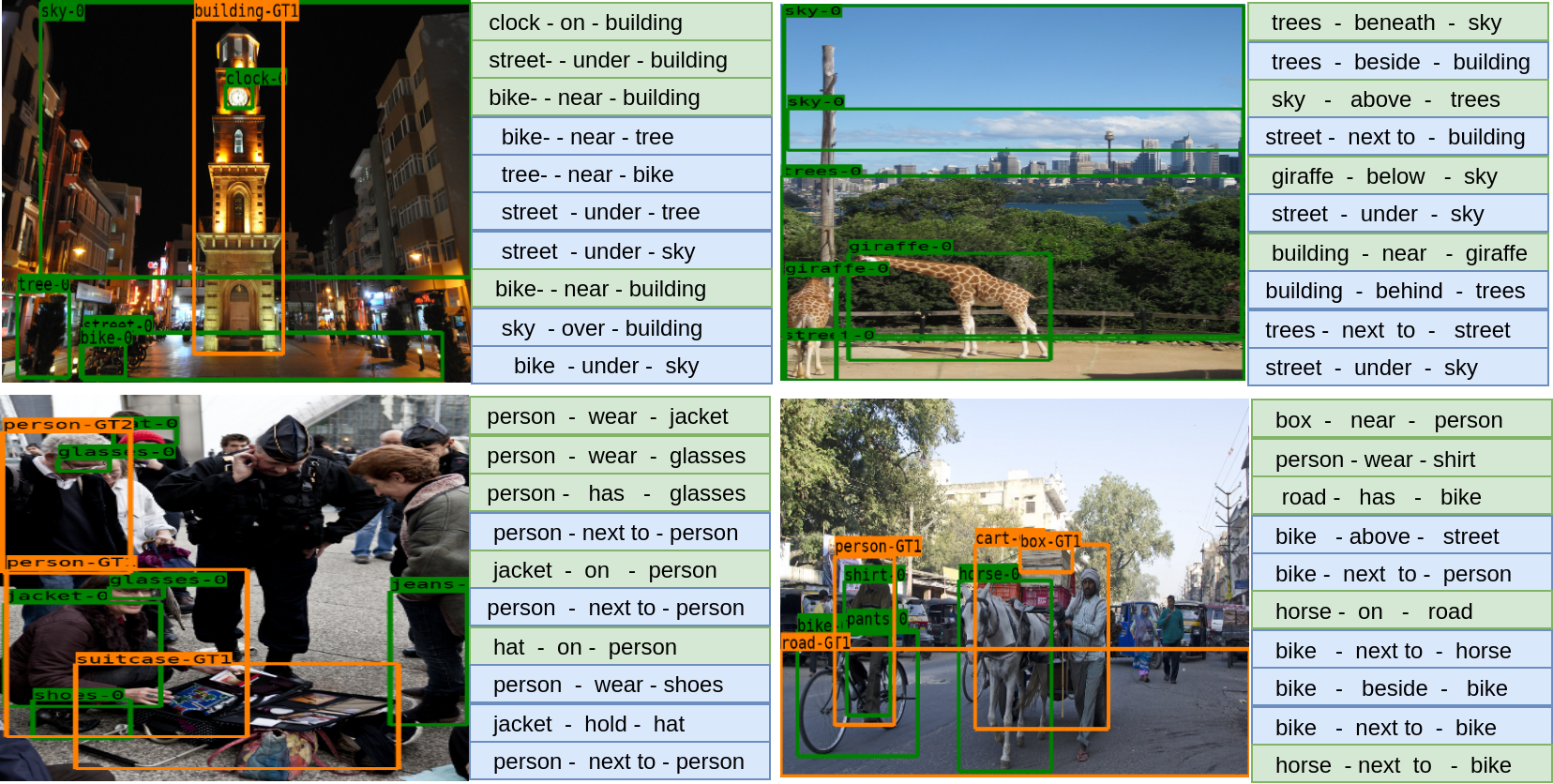}
\caption{Most probable results predicted by our model. Green shade indicates a match with ground truth labels. Matched results indicate strong presence of multimodality and the triplets generated that do not match are mostly reasonable.}
\label{fig:qual_res}
\end{figure}
\begin{figure}[t]
\centering
\includegraphics[width=0.9\columnwidth]{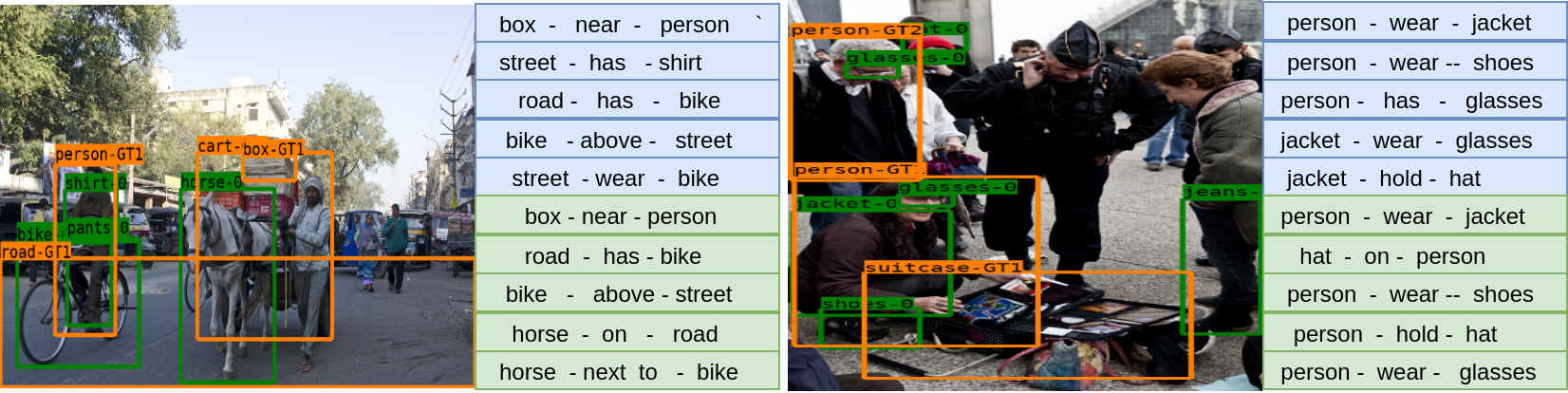}
\caption{Results of model with prior indicate it helps in removing spurious triplets. \colorbox{blue}{without prior}\colorbox{green}{with prior}}
\label{fig:qual_res_2}
\end{figure}
The ablation results are shown in Table~\ref{tab:ablations}. Both the datasets show significant improvement when the rank is increased from rank-1 to rank-5. Note that rank-1 tensor decomposition is equivalent to independent distributions of triplets and rank-5 to a mixture of 5 components. From rank-1 to rank-5, recall@50 score with multiple predictions per box pair increases by 5\% points in the VG dataset and by 2\% points in the VRD dataset. This shows that tensor decomposition structure is better able to capture the multimodal triplet distribution. Clearly, gain in score with multiple predictions ($k>1$) is better than with single prediction ($k=1$) and with recall@100 is better than recall@50 across datasets. This result is in line with our motivation of optimizing the model for multiple relations instead of one. With a rank-1 independent distribution, the model is optimized for a single top-prediction, hence recall@50 with $k=1$ has similar score across ranks 1 to 5. With mixture distribution the model is optimized for a set of valid predictions, hence recall@50/100 score for $k>1$ increases substantially with increasing rank. As we further increase the mixing components, there is a small reduction in score. From this, we can infer that most conditional triplet distributions have a small number of modes and increasing the rank further just increases number of parameters in the model. Further including the $X_{sel}$ variable and dataset prior to the Rank-5 model improves the score substantially supporting our assumption of strong bias towards a small set of relationships. 
%This particularly seen in VRD dataset evalution on relationship detection. As was shown in ~\cite{jae2018tensorize,baier2017improving}, there is much information to be extracted from the label counts of the traning data. Our natural incorporation of this dataset factor helps in reaching the best state-of-the art score on both VRD and VG datasets. 
\begin{table}
\begin{center}
\scriptsize
\renewrobustcmd{\bfseries}{\fontseries{b}\selectfont}
\renewrobustcmd{\boldmath}{}
\begin{tabular}{c|c|ccccc|}
		\hline 
		\multirow{3}{*}{Dataset} & \multirow{3}{*}{Ablation model} & \multicolumn{4}{c}{Phrase classification} \\
		& & \multicolumn{2}{c}{mult preds (free k)} 
		& \multicolumn{2}{c}{single pred (k=1)} \\
		%\cline{2-6}
		& & R@50 & R@100 & R@50 & R@100  \\
		\hline
		\multirow{7}{*}{VG150} 
		& Rank-1  & 37.35 & 44.49 & 32.16 & 34.30 \\ 
		& Rank-2  &  42.53 &  47.52 &  33.64 & 35.26\\
	    & Rank-3 & 41.89 & 48.35 & \bfseries 33.89 & 35.31\\
		& Rank-4  & 41.74  & 48.22 & 33.29 & 35.14\\ 
		& Rank-5  & \bfseries 42.71 & \bfseries 48.95 & 32.21 & \bfseries 35.35\\  
		\cline{2-6}
		& w/ $X_{sel}$ &  46.25 &  53.05 & 34.57 &  36.57\\
		& w/ $X_{sel}$ \& prior & \bfseries 47.54  & \bfseries 54.69 & \bfseries 35.60 & \bfseries 37.68\\ 
		\hline 
		
		\multirow{7}{*}{VRD} 
		& Rank-1  & 39.01 & 46.3 & 31.12 & 33.04 \\ 
		& Rank-2  & \bfseries 41.29 & \bfseries 51.58 & \bfseries 33.43 & \bfseries 35.16\\ 
	    & Rank-3 & 40.28 & 49.5 & 32.46 & 34.96\\ 
		& Rank-4  & 39.63  & 49.64 & 32.03 & 33.52\\
		& Rank-5  & 40.47 &  50.75 & 32.59 & 34.06\\
		\cline{2-6}
		& w/ $X_{sel}$ &  42.89 &  52.47 & 33.51 &  35.78\\
		& w/ $X_{sel}$ \& prior & \bfseries 44.02  & \bfseries 53.99 & \bfseries 34.14 & \bfseries 36.07\\
		\hline
\end{tabular}
\end{center}
\caption{Ablation results on the task of phrase classification.}
\label{tab:ablations}
\end{table}

\subsection{Qualitative Results:}
Figure~\ref{fig:qual_res} shows some of the qualitative results of our model. From the overlap between ground-truth and predicted labels, it can be seen that the conditional distribution is at least bimodal if not tri-modal. Also, the triplets that are generated by the models but are not annotated are mostly reasonable. 

In Figure~\ref{fig:qual_res_2}, we visualize results without prior multiplication. We see multiple cases of erroneous phrases such as `street-has-shirt', `street-wear-bike', and `jacket-wear-glasses' that are unlikely to appear in common usage. Interestingly, such spurious triplets are pushed down from the top of the output lists after multiplication with prior. However, non-spurious triplets that do not appear in the training set may also be pushed down; using external language datasets may improve performance by providing improved usage prior.
%This effect suggests that the use of external language datasets will likely be able to further reduce these types of errors.
%As we are not training the main model on negative instances, the top prediction results seem mostly correct though they may not match the ground truth. 

\section{Conclusion}
We observe that the conditional distribution of relation triplets given input bounding box pair in relation detection tasks is often multimodal. We propose use of mixture of rank-1 tensors for modeling the conditional distribution. This enables the model to capture multimodal properties of the distribution with a reasonably small number of model parameters while being efficiently trainable. We further model the generative labeling process to handle missing annotations and remove spurious triplets with principled incorporation of dataset prior. We show that each of these improve performance on the task of visual relationship recognition. Further improvements may include a language prior from external datasets and with our tensor-decomposition model, it should be possible to do graph inference with higher-order triplet potential over all the boxes in the image.

\section*{Acknowledgements}
This work is supported by NUS AcRF Tier 1 grant R-252-000-639-114.

\bibliography{AAAI-DuptyM.4633.bib}
\bibliographystyle{aaai}
%We incorporate triplet potential involving subject, predicate and object variables into our model. Representing the triplet potential exactly requires scoring all possible triplet combinations. For large number of objects and predicates, learning a parameterized function mapping from the triplet image feature space to the a very large dimensional potential function space becomes intractable given the limited amount of data available for learning such a mapping. Our novel method of representing the high order triplet potentials as low rank tensors enables us to efficiently learn this function from limited data and consequently run inference with high order potentials included. 

%This technique exponentially reduces the number of parameters required to learn such a high order potential function and hence making it possible to learn this factor from less data. Apart from reducing the parameters, certain operations like marginalization can be performed efficiently when we have such a representation, which is one of the operation in running the belief propagation algorithm for inference as was shown in \cite{wrigley2017tensor}. Consequently, approximate inference is possible even when we have high order potentials in the graph.

\onecolumn
\section{Visual Relationship Detection with Low Rank Non-Negative Tensor Decomposition \\Supplementary material for Submission ID: 4633}

\subsection{Derivative of Cross-Entropy Loss with non-negative tensor decomposition}

In this section, we derive the derivative of the cross-entropy loss when the model distribution is represented as a mixture of rank-1 tensors with respect to the vectors of the rank-1 tensors output by the neural network. To keep up with conventional notations, we follow notations in ~\cite{sadowski2016notes}, which describes gradient computation of a single layer neural network. 

Let  $\psi_c(X_s, X_p, X_o| B_s, B_o, I)$ represent the tensor of interest, a function of 3 variables $X_s, X_p, X_o$ with subscript $s,p,o$ representing subject, predicate and object. Extension to further number of variables is straightforward. For a rank-$R$ approximation, the network outputs $3R$ vectors, $R$ vectors each for subject, predicate and object. If $\phi_{rs}$ represents $r^{th}$ vector for variable $X_s$, $\psi_c$ can be represented as
\begin{dmath}
	\psi_c(X_s, X_p, X_o| B_s, B_o, I) =  \sum_{r=1}^{R}  \phi_{rs}(X_s|B_s, B_o, I)\otimes \phi_{rp}(X_p|B_s, B_o, I)\otimes\phi_{ro}(X_o|B_s, B_o, I)).
	\label{tensor_potential}
\end{dmath}
where $\otimes$ is the outer product operator. For notational convenience, we drop the conditioning on $B_s, B_o$ and $I$ and represent $\psi_c$ as
\begin{dmath}
	\psi_c(X_s,X_p,X_o) = \sum_{r=1}^{R} \phi_{rs}(X_s) \otimes \phi_{rp}(X_p) \otimes \phi_{ro}(X_o)
\end{dmath}
Let $s$ represent the final layer output of the network. We exponentiate $s$ to make it non-negative i.e. $\phi_{ra}(X_a=i) = e^{s_{a,r}^i}$ for $a\in \{s,p,o\}$. If each of the $R$ vectors are indexed by $i,j,k$ for variables $X_s, X_p$ and $X_o$ respectively, then value of $\psi_c$ at index $(i,j,k)$ is
\begin{dmath}
	\psi_c(X_s=i,X_p=j,X_o=k) = \sum_{r=1}^{R} e^{s_{s,r}^i} \cdot e^{s_{p,r}^j} \cdot e^{s_{o,r}^k}
\end{dmath}
We then normalize $\psi_c$ to represent it as a probability distribution $y$ with $y^{i,j,k}$ given by

\begin{dmath}
	y^{i,j,k} = \frac{\sum_{r=1}^{R} e^{s_{s,r}^i + s_{p,r}^j + s_{o,r}^k}}{\sum_{l,m,n} \sum_{r=1}^{R} e^{s_{s,r}^l + s_{p,r}^m + s_{o,r}^n} }
\end{dmath}
Let $Z$ represent the partition function which can be efficiently computed by pushing the sum $\sum_{l,m,n}$ inside,

\begin{dmath}
	Z = \sum_{l,m,n} \sum_{r=1}^{R} e^{s_{s,r}^l + s_{p,r}^m + s_{o,r}^n} 
	= \sum_{r=1}^{R} \sum_l e^{s_{s,r}^l} \sum_m  e^{s_{p,r}^m} \sum_n e^{s_{o,r}^n}.  
\end{dmath}
If $t$ represents the target tensor and $y$ the output tensor, the cross-entropy loss function is
\begin{dmath}
	\label{eq:main}
	L = -\sum_{i,j,k} t^{i,j,k}\log(y^{i,j,k}).
\end{dmath}

\iffalse
If $t$ represent the target one-hot tensor (tensor that is zero everywhere, except at a single tuple index which has value 1, representing the indicator function of the tuple) and $y$ the output tensor, the cross-entropy loss function is

Each of these  $R$ vectors are indexed by $i,j,k$ for subject, predicate and object categories respectively.
\begin{dmath}
	y_{i,j,k} = \frac{\sum_{r=1}^{R} e^{s_{1,r}^i} \cdot e^{s_{2,r}^j} \cdot e^{s_{3,r}^k}}{\sum_{l,m,n} \sum_{r=1}^{R} e^{s_{1,r}^l} \cdot e^{s_{2,r}^m} \cdot e^{s_{3,r}^n} }
\end{dmath}
\fi 
Computing gradients of loss $L$ w.r.t $s_{s,r'}^i$, where $s_{s,r'}^i$ is the $i^{th}$ network output for variable $X_s$ and $r'$ refers to one of the $R$ vectors. 
\iffalse
\begin{equation}
	\label{eq:main}
	\frac{\partial L}{\partial s_{s,r'}^i} = \sum_{l,m,n}\frac{\partial L}{\partial y_{l,m,n}} \frac{\partial y_{l,m,n}}{\partial s_{s,r'}^i}
\end{equation}
\fi
\begin{dmath}
	\frac{\partial L}{\partial y^{i,j,k}} = -\frac{t^{i,j,k}}{y^{i,j,k}} 
\end{dmath}

For $l=i$:
\begin{dmath}
	\frac{\partial y^{i,j,k}}{\partial s_{s,r'}^l} = 
	\frac{e^{s_{s,r'}^i + s_{p,r'}^j + s_{o,r'}^k} Z - \frac{\partial Z}{\partial s_{s,r'}^l} \sum_{r=1}^{R} e^{s_{s,r}^i + s_{p,r}^j + s_{o,r}^k} }{Z^2}
\end{dmath}
\begin{dmath}
	\frac{\partial Z}{\partial s_{s,r'}^l} = e^{s_{s,r'}^i}\sum_{m,n}e^{s_{p,r'}^m + s_{o,r'}^n}  
\end{dmath}
\begin{dmath}
	\label{eqn:leqi}
	\frac{\partial y^{i,j,k}}{\partial s_{s,r'}^l} = \frac{e^{s_{s,r'}^i + s_{p,r'}^j + s_{o,r'}^k}}{Z} -  y^{i,j,k} \frac{e^{s_{s,r'}^i}\sum_{m,n}e^{s_{p,r'}^m + s_{o,r'}^n}}{Z}
\end{dmath}
For $l\neq i$:
\iffalse
\begin{dmath}
	\frac{\partial y^{i,j,k}}{\partial s_{s,r'}^l} =
	-  \frac{ e^{s_{s,r'}^l}\sum_{m,n}e^{s_{p,r'}^m + s_{o,r'}^n} \sum_{r=1}^{R} e^{s_{s,r}^i + s_{p,r}^j + s_{o,r}^k}}{Z^2}
\end{dmath}
\fi 
\begin{dmath}
	\label{eqn:lneqi}
	\frac{\partial y^{i,j,k}}{\partial s_{s,r'}^l} = - \frac{y^{i,j,k} e^{s_{s,r'}^l}\sum_{m,n}e^{s_{p,r'}^m + s_{o,r'}^n}}{Z}
\end{dmath}

From (\ref{eq:main}) we have,
\begin{dmath}
	\frac{\partial L}{\partial s_{s,r'}^i} = \sum_{l,m,n}\frac{\partial L}{\partial y^{l,m,n}} \frac{\partial y^{l,m,n}}{\partial s_{s,r'}^l}
\end{dmath}
\begin{dmath}
	\frac{\partial L}{\partial s_{s,r'}^i} = \sum_{m,n}\frac{\partial L}{\partial y^{i,m,n}} \frac{\partial y^{i,m,n}}{\partial s_{s,r'}^i} + \sum_{l\neq i, m,n} \frac{\partial L}{\partial y^{l,m,n}} \frac{\partial y^{l,m,n}}{\partial s_{s,r'}^l}
\end{dmath}
From (\ref{eqn:leqi}) and (\ref{eqn:lneqi}), we have
\begin{dmath}
	\frac{\partial L}{\partial s_{s,r'}^i} = \sum_{m,n} -t^{i,m,n} \frac{ \frac{e^{s_{s,r'}^i + s_{p,r'}^m + s_{o,r'}^n}}{y^{i,m,n}} - e^{s_{s,r'}^i}\sum_{m',n'}e^{s_{p,r'}^{m'} + s_{o,r'}^{n'}}}{Z} + \sum_{l\neq i,m,n} t^{l,m,n} \frac{ e^{s_{s,r'}^i}\sum_{m',n'}e^{s_{p,r'}^{m'} + s_{o,r'}^{n'}}}{Z}
\end{dmath}
\begin{dmath}
	=\sum_{m,n} -t^{i,m,n} \frac{e^{s_{s,r'}^i + s_{p,r'}^m + s_{o,r'}^n}}{\sum_{r=1}^{R} e^{s_{s,r}^i + s_{p,r}^m + s_{o,r}^n}}  +  \sum_{l,m,n} \frac{t^{l,m,n}e^{s_{s,r'}^i}\sum_{m',n'}e^{s_{p,r'}^{m'} + s_{o,r'}^{n'}}}{Z} 
\end{dmath}
\begin{dmath}
	=\sum_{m,n} -t^{i,m,n} \frac{e^{s_{s,r'}^i + s_{p,r'}^m + s_{o,r'}^n}}{\sum_{r=1}^{R} e^{s_{s,r}^i + s_{p,r}^m + s_{o,r}^n}}  +  \frac{e^{s_{s,r'}^i}\sum_{m',n'}e^{s_{p,r'}^{m'} + s_{o,r'}^{n'}}}{Z}\sum_{l,m,n} t^{l,m,n}
\end{dmath}

\begin{dmath}
	=\frac{e^{s_{s,r'}^i}\sum_{m',n'}e^{s_{p,r'}^{m'} + s_{o,r'}^{n'}}}{Z} + \sum_{m,n} -t^{i,m,n} \frac{e^{s_{s,r'}^i + s_{p,r'}^m + s_{o,r'}^n}}{\sum_{r=1}^{R} e^{s_{s,r}^i + s_{p,r}^m + s_{o,r}^n}}
\end{dmath}
For Sparse Cross-Entropy loss, the target tensor $t$ is one-hot encoded (tensor that is zero everywhere, except at a single tuple index which has value 1, representing the indicator function of the tuple). For an observed tuple $(i',j',k')$, the derivative of the Loss $L$ with respect to $s_{s,r'}^i$ is
\[ 
\frac{\partial L}{\partial s_{s,r'}^i} =
\begin{cases}
\frac{e^{s_{s,r'}^i}\sum_{m'}e^{s_{p,r'}^{m'}}  \sum_{n'}e^{s_{o,r'}^{n'}}}{Z} & i\neq i' \\
\\
\frac{e^{s_{s,r'}^i}\sum_{m'}e^{s_{p,r'}^{m'}}  \sum_{n'}e^{s_{o,r'}^{n'}}}{Z} \\ - 
\frac{e^{s_{s,r'}^i + s_{p,r'}^{j'} + s_{o,r'}^{k'}}}{\sum_{r=1}^{R} e^{s_{s,r}^i + s_{p,r}^{j'} + s_{o,r}^{k'}}}
& i=i'
\end{cases}
\]

%%%%%%%%%%%%%%%%%%%%%%%%%%%%%%%%%%%%%%%%%%%%%%%%%%%
%%%%%%%%%%%%%%%%%%%%%%%%%%%%%%%%%%%%%%%%%%%%%%%%%%%

\end{document}